\documentclass[sigconf]{acmart}
\AtBeginDocument{%
  \providecommand\BibTeX{{%
    \normalfont B\kern-0.5em{\scshape i\kern-0.25em b}\kern-0.8em\TeX}}}

\copyrightyear{2023}
\acmYear{2023}
\setcopyright{acmlicensed}\acmConference[MM '23]{Proceedings of the 31st ACM International Conference on Multimedia}{October 29-November 3, 2023}{Ottawa, ON, Canada}
\acmBooktitle{Proceedings of the 31st ACM International Conference on Multimedia (MM '23), October 29-November 3, 2023, Ottawa, ON, Canada}
\acmPrice{15.00}
\acmDOI{10.1145/3581783.3612871}
\acmISBN{979-8-4007-0108-5/23/10}

\begin{document}

\title{Query-aware Long Video Localization and Relation Discrimination for Deep Video Understanding}

\author{Yuanxing Xu}
\email{xyx@bupt.edu.cn}
\orcid{0009-0003-5039-9937}
\affiliation{%
  \institution{Beijing University of Posts and Telecommunications}
  \state{Beijing}
  \country{China}
}

\author{Yuting Wei}
\email{yuting_wei@bupt.edu.cn}
\orcid{0000-0002-3971-708X}
\affiliation{%
  \institution{Beijing University of Posts and Telecommunications}
  \state{Beijing}
  \country{China}
}

\author{Bin Wu}
\authornote{Corresponding author.}
\email{wubin@bupt.edu.cn}
\orcid{0000-0002-7112-126X}
\affiliation{%
  \institution{Beijing University of Posts and Telecommunications}
  \state{Beijing}
  \country{China}
}

\renewcommand{\shortauthors}{Yuanxing Xu, Yuting Wei, \& Bin Wu}


\begin{abstract}
  The surge in video and social media content underscores the need for a deeper understanding of multimedia data. Most of the existing mature video understanding techniques perform well with short formats and content that requires only shallow understanding, but do not perform well with long format videos that require deep understanding and reasoning. Deep Video Understanding (DVU) Challenge aims to push the boundaries of multimodal extraction, fusion, and analytics to address the problem of holistically analyzing long videos and extract useful knowledge to solve different types of queries. This paper introduces a query-aware method for long video localization and relation discrimination, leveraging an image-language pretrained model. This model adeptly selects frames pertinent to queries, obviating the need for a complete movie-level knowledge graph. Our approach achieved first and fourth positions for two groups of movie-level queries. Sufficient experiments and final rankings demonstrate its effectiveness and robustness.
\end{abstract}





\maketitle

\section{Introduction}

With the explosion of video content on social media, deep understanding of long videos has become critical in areas such as movies, TV shows, and online education. However, it remains a challenge to effectively extract key data from these videos. Deep Video Understanding (DVU) Challenge 2023 focuses on the use of knowledge graphs to understand entity relations in long videos. Long videos pose challenges in processing due to their long duration, multiple scenes and complex plots.

Existing video processing methods \cite{lee2019self,gao2018motion,li2019beyond,jang2017tgif,jiang2020reasoning,li2022invariant,park2021bridge} mainly target short format or single-scene videos, relying on a superficial understanding. They often underperform when handling lengthy and complex videos that demand deeper comprehension and reasoning. Compared to short videos, long ones have significantly more content variety and complexity. Since short videos typically encompass single or few scenes, their intricacy is much lower than long videos. Thus, the effectiveness of existing methods for short videos doesn't seamlessly translate to long video processing.

To address the issues of the DVU Challenge 2023, we propose an innovative approach for processing long videos, grounded on an image-language pretrained model and optimizing query-aware features. Unlike traditional techniques, our method avoids processing all multimodal data at once but refines the search scope to precisely locate query-related information from vast data. Initially, visual features are extracted from the videos based on given entity images, their names, and types. Then, we employ a frame selector designed to identify frames closely associated with the entity in the query from sampled frames. Subsequently, leveraging the relations in the query, the features of the chosen frames and the subtitles from their time segments are utilized to calculate relation scores between entities. Ultimately, the entity with the highest cumulative relation scores in the selected frames is chosen as the answer. The strength of our method lies in its capacity to swiftly and accurately pinpoint relevant data in extensive long videos without necessitating the construction of an exhaustive movie-level knowledge graph, facilitating profound video comprehension. The results of our main and robustness experiments on the validation and test sets demonstrate the effectiveness and robustness of our approach.

In summary, this paper targets the difficulty of long video understanding and proposes a new query-aware method for processing long videos. With this method, we can not only solve the challenges faced by existing long video processing methods but also enhance the depth of understanding of video content, providing possibilities for further development of video social media.

\section{Related Work}

\textbf{Video Question Answering.}
Video Question Answering (VideoQA) is a typical example of video-language tasks. Motion-appearance memory \cite{lee2019self,gao2018motion,li2019beyond}, cross-modal attention \cite{jang2017tgif,li2019beyond}, and graph-based methods \cite{jiang2020reasoning,li2022invariant,park2021bridge}, have all made some progress. Transformer-based models \cite{li2020hero,xu2021videoclip,seo2021look,yang2021just}, have had great success on visual language tasks in recent years, achieving encouraging performance on many benchmark datasets. However, most of the videos in these datasets are only a few seconds to tens of seconds obtained from the web, with a single scene and simple plot, and most of the problems set are descriptive and localized. Some work \cite{gao2021env,grunde2021agqa,wu2021star,xiao2021next} has also noted this, developing benchmark datasets that emphasize visual relational reasoning, increasing the length of the videos, and setting up problems that require reasoning. However this does not significantly compensate for the above shortcomings. The analysis and deep understanding of movie-level long videos is still in the early stages of exploration.

\textbf{Vision-Language Pretrained Models.}
With the great success of the vision-language pretrained models, some work \cite{cheng2022tallformer,wu2021towards} has directly used the VideoQA task as a downstream task on which to fine-tune the pretrained model. The image-language pretrained model \cite{driess2023palm,wang2022ofa,lei2021less,li2022blip,yao2021filip} has more advances than the video-language pretrained model \cite{wang2022internvideo,zellers2022merlot,ma2023temporal,xu2023mplug,wang2023all}. In this paper, our work builds on two of the current state-of-the-art image-language pretrained models \cite{li2022grounded,li2023blip} for entity detection and question answering, respectively.

\textbf{Long Video Modeling.}
The great progress in short video understanding has motivated some pioneering work \cite{cheng2022tallformer,wu2021towards} to focus on long video modeling using action recognition or localization tasks. \cite{cheng2022tallformer} proposes short-term feature extraction and long-term memory mechanisms to reduce the processing of redundant video frames during training. \cite{islam2022long} introduces structured multi-scale temporal decoder for self-attention. \cite{lin2022eclipse} suggests replacing the corresponding video with information-rich audio. They all work to reduce the computational effort of long video modeling to improve the performance. In this paper, we use query bootstrapping to localize to some frames that are most relevant to the query from the perspective of human beings performing deep comprehension of long videos and answering questions for the same purpose.

\section{Method}

\begin{figure*}[htbp]
  \centering
  \includegraphics[width=0.8\linewidth]{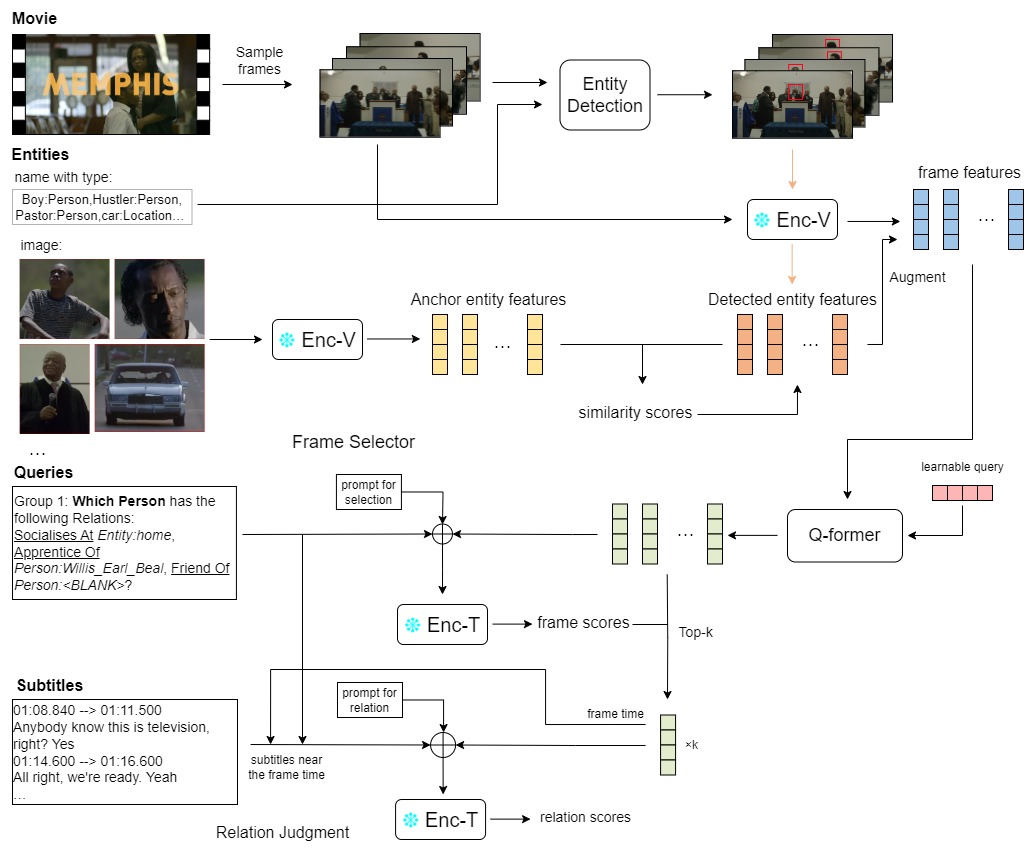}
  \caption{Pipeline of our method.}
    \label{Pipeline}
\end{figure*}

Our method's pipeline is depicted in Figure \ref{Pipeline}. Initially, we extract visual features from a long video using given entity images, names, and types. Next, a frame selector identifies relevant frames from a subset based on query-mentioned entities. Then, using these frames' features and their subtitles, we compute relation scores among entities. The answer is derived from the highest aggregated relation scores in the selected frames.

\subsection{Entity Detection and Feature Representation}

We uniformly sample frames with their time in the movie at a rate of $p$ seconds per frame. Subsequently, the sampled frames, the entity names and types of the movie are fed into the object detector GLIP \cite{li2022grounded} to localize the entities appearing in the frames, as shown at the top of Figure \ref{Pipeline}. Both the detected entity images and the given entity images are fed into ViT \cite{dosovitskiy2020image} to extract visual features. Next, in order to know the names of the detected entities, we calculate the cosine similarity between the detected entities and their cosines, using the given entity images as anchors, to determine the names of the entities: 
\begin{equation}
    sim_{i,j} = \psi \left( f_{i}^{anc}, f_{j}^{t} \right)
\end{equation}
where $f_{i}^{anc}$ represents the feature of the $i$-th anchor image, $f_{j}^{t}$ represents the $j$-th entity feature of the $t$-th frame, and $\psi$ is the cosine similarity function. In this way, we get the features of all the entities in the entity set in the sampled frames.

In addition, we apply self-attention pooling \cite{lee2019self} to the features of the entities in the frame in order to make the aggregated features strengthen the features of the frame:
\begin{equation}
    f_{pool}^{t} = \sum_{i=1}^{*} \alpha_{i}^{t} f_{i}^{t}, \quad \alpha=\sigma \left( \phi_{W_{p}} \left( F \right) \right)
\end{equation}
\begin{equation}
    f_{aug}^{t} = \phi_{W_{a}} \left( \left[ f_{i}^{t};f_{pool}^{t}  \right] \right)
\end{equation}
where $W_{p}$ and $W_{a}$ are the linear mapping weights. The purpose of this operation is that in the subsequent part of the frame selector, when an entity mentioned in the query is present in the frame, it can produce a stronger response, while avoiding the possibility that the frame will be less likely to be selected due to the excessive influence of the background information of the frame.

\subsection{Frame Selection} \label{Frame Selection}

The sampled frames extracted in the previous section are relatively dense and contain a lot of redundant information that is not relevant to the query. For this reason, we design a frame selector, which uses BLIP-2 \cite{li2023blip} as a skeleton, to first characterize the sampled frames with a learnable query $q$ (considered as parameters) as a Q-former input:
\begin{equation}
    f_{mm}^{t} = \mathrm{Qformer} \left( \left[ f_{aug}^{t}; q \right] \right)
\end{equation}
This further extracts semantically rich visual information and allows it to break through the barrier between visual and textual modalities. This visual information is then fed into Flan-T5 along with the entity names in the query and the prompt used to select the frames, which is used to compute the score $s_{f}^{t}$ for each frame $t$:
\begin{equation} \label{eq:FrameSel}
    s_{f}^{t} = \mathrm{FlanT5} \left( \left[ f_{aug}^{t}; entity; prompt_{f} \right] \right)
\end{equation}
in which we use the question "Does the frame contain this entity?" as the prompt for frame selection in our work. The $K$ frames with the highest scores are selected since they are considered as frames that are strongly related to the entities involved in the query.

\subsection{Relation Discrimination}

In Section \ref{Frame Selection}, we filtered out the $K$ frames that are most relevant to the entity e mentioned in the query, since we believe that two entities interacting in a video usually appear in the same frame. Therefore there is a high probability that the correct answer is among the other entities in these $K$ frames. 

First, the time at which these $K$ frames are in the movie helps us locate the subtitles in their vicinity. This does so on the one hand by picking out the part of the subtitle that is most likely to contribute to the answer from the whole movie's subtitle, avoiding the need to analyze the whole movie's subtitle, and on the other hand, the subtitle of the time period in which the frame is located helps to supplement some of the information of the frames that have not been sampled. Similar to Eqn. \ref{eq:FrameSel}, we feed the $K$ frame features and their time-neighboring subtitles, the relation $r$ mentioned in the query with the prompt used for relation identification into Flan-T5, and compute the relation scores of $e$'s relations with other entities in the frame with respect to $r$:
\begin{equation}
    s_{r}^{k} = \mathrm{FlanT5} \left( \left[ f_{mm}^{k}; query; prompt_{r} \right] \right), \quad k=1,2,\ldots,K
\end{equation}
where we use the question "Do other entities in the frame have this relation with this entity?" as the prompt for relation discrimination.

\subsection{Query Decomposition and Answer Prediction}

Since the query contains multiple conditions (e.g. the example query in Figure \ref{Pipeline} has three conditions: Socialises At Entity:home, Apprentice Of Person:Willis\_Earl\_Beal and Friend Of Person:<BLANK>). It is more difficult to use all the conditions at once for frame selection and relation recognition and the selected frames may be inaccurate. For this reason, we divide the query into multiple subqueries with individual conditions, which are fed into the frame selector and the relation score calculation module, respectively. In this way, each subquery will get the relation score of $e$ with a set of entities. Eventually, the scores of all corresponding entities are summed up and the entity with the highest score is selected as the predicted answer $a^{*}$:
\begin{equation}
    a^{*} = \arg\max \limits_{e \in \mathcal{E}} \sum_{k=1}^{K}  s_{r_e}^{k}
\end{equation}

\subsection{Modifications for Question Answering} 

For movie-level Group 1, questions have a relatively fixed format, presenting clear subject-predicate-object triples. This design aids in frame selection and relation discrimination by using objects and predicates. However, Group 2 questions, specifically QA, are different. They don't provide similar triples and are more diverse. Our originally proposed pipeline, designed with Group 1 in mind, isn't readily applicable to the generalized QA tasks. Thus, we replace the relation discrimination module with a QA module but retained the same structure. Furthermore, we input the full question and its options into Flan-T5 in both modules and set a particular prompt for the question-answer module to ensure its accuracy.

\section{Experiments}

In this section, we first introduce the dataset used in DVU Challenge 2023 and our experimental settings. We then present some experimental results on the test dataset and analyze them, including experiments on the original dataset as well as robustness experiments on the dataset with visual noise perturbations added.

\subsection{Dataset and Experimental Settings}

The full Deep Video Understanding training set of 19 movies and total duration of ~25 hours. This training set has been annotated by human assessors and final ground truth, both at the overall movie level (\textbf{Ontology of relations}, \textbf{entities}, actions \& events, \textbf{Knowledge Graph}, and \textbf{names and images of all main characters}) and the individual scene level. \textbf{Bold items} represent the data we used. The test set is comprised of 5 movies licensed from KinoLorberEdu platform. We used the 5 movies that served as the 2022 test set as a validation set to examine the performance prior to the release of ground truth from the test set.

During the experiments, in order to evaluate the performance of both the Fill-in-the-graph-space and QA tasks, we choose accuracy as the metric. The difference is that for the QA task, we directly use \textbf{Acc} as the metric, whereas for the Fill-in-the-graph-space task, we map the relation scores to the confidence of each candidate entity (summed up to 100) and sort them in descending order. As a result, we use \textbf{Acc@n} as the metric, which represents the probability that the correct answer appears in the first n answers.

\subsection{Experimental Results and Analysis}

\begin{table}
  \caption{Test Acc (\%) of Fill-in-the-graph-space task. Before and after the slash are the results of the main task and the robustness task, respectively. Same below.}
  \label{tab:test_FG}
  \begin{tabular}{ccccc}
    \toprule
    Movie & No.queries & Acc@1 & Acc@2 & Acc@3  \\
    \midrule
    Archipelago & 4 & 50 / 50 & 75 / 75 & 100 / 100 \\
    Bonneville & 4 & 50 / 25 & 75 / 50 & 75 / 75 \\
    Heart Machine & 4 & 75 / 75 & 100 / 75 & 100 / 100 \\
    Little Rock & 4 & 50 / 50 & 50 / 50 & 75 / 75 \\
    Memphis & 4 & 75 / 50 & 75 / 75 & 75 / 75 \\
    \hline
    Total & 20 & 60 / 50 & 75 / 65 & 85 / 85 \\
  \bottomrule
\end{tabular}
\end{table}

\begin{table}
  \caption{Test Acc (\%) of QA task.}
  \label{tab:test_QA}
  \begin{tabular}{cccc}
    \toprule
    Movie & No. queries & No. correct & Acc\\
    \midrule
    Archipelago & 27 & 10 / 9 & 37.04 / 33.33 \\
    Bonneville & 46 & 16 / 16 & 34.78 / 34.78 \\
    Heart Machine & 26 & 13 / 10 & 50 / 38.46 \\
    Little Rock & 30 & 8 / 7 & 26.67 / 23.33 \\
    Memphis & 22 & 10 / 10 & 45.45 / 45.45 \\
    \hline
    Total & 151 & 57 / 52 & 37.75 / 34.44 \\
  \bottomrule
\end{tabular}
\end{table}

The specific performance of the Fill-in-the-graph-space task is shown in Table \ref{tab:test_FG}, where our method achieves no less than 60\% accuracy on both the validation and test sets (Acc@1). As can be seen from Acc@2 and Acc@3, for most of the queries that are not answered correctly, the correct answers also rank very high in the set of candidate answers given. Our method achieves first place in the test set for this task. This demonstrates our method exhibits significant performance on long video localization and relation discrimination tasks. By accurately filtering out frames that are strongly correlated with the query, our method is able to effectively discriminate relations. 

Our method also achieves good results in the QA task, as shown in Table \ref{tab:test_QA}, and ranks fourth in the test set for that task. However, their accuracy is not as high as that of the Fill-in-the-graph-space task, most likely due to the fact that the queries in the QA task have a wider range of forms, and the design of the relation discrimination in our method is more refined than that of QA.

In addition, compared to the validation set, the performance of the test set slightly decreases on both tasks, especially on the question and answer task. This may be due to the fact that the movie plots and entity relations in the test set are more complex and have more distractors. On the other hand, most of the queries in the validation set were relational (e.g., "What is the relation/connection from A to B?") , which is essentially just another form of query to the Fill-in-the-graph-space task. However, the test set adds more non-relational queries, which require temporal and causal reasoning, such as "what", "who", and "why" types of questions, which require models with stronger reasoning capabilities.

\subsection{Experiments on Robustness}

We also performed an advanced robustness subtask in which the movies in the dataset were subjected to various noise perturbations. Considering that our method does not use audio, the robustness assessment is performed only on datasets that are affected by visual noise. As shown in Tables \ref{tab:test_FG} and \ref{tab:test_QA}, the performance of our method is minimally affected by visual noise perturbations, which highlights its excellent robustness. This is partly attributed to the pretraining of the visual model used on a variety of datasets filled with real noise and perturbations. Such pretraining ensures that our frame selector is largely unaffected. At the same time, our approach is primarily query-driven, bolstered by captioning data, and has little reliance on visual cues. Thus, even when the quality of the extracted visual features is not as high, the model can still utilize the textual data and prompts to make relatively accurate inferences.

\section{Conclusion}

In this paper, we propose a new approach for query-aware long video localization and relation discrimination in movie-level long videos. By utilizing an image-language pretrained model, we successfully filter out frames that are highly correlated with a given query, allowing us to discriminate relations between corresponding entities without the need to construct an explicit movie-level knowledge graph. Our approach was subjected to main and robustness experiments on two movie-level tasks, both of which yielded promising results, demonstrating its effectiveness and robustness in addressing the challenges associated with large-scale video analytics.

\begin{acks}
This work is supported by the NSFC-General Technology Basic Research Joint Funds (grant no. U1936220) and the National Natural Science Foundation of China (grant no. 61972047).
\end{acks}

\bibliographystyle{ACM-Reference-Format}
\balance
\bibliography{sample-base}


\end{document}